\documentclass[10pt,twocolumn,letterpaper]{article}

\usepackage{cvpr}              

\makeatletter
\renewcommand\paragraph{\@startsection{paragraph}{4}{\z@}%
  {4pt \@plus 1pt \@minus .2ex}%
  {-1em}%
  {\normalfont\normalsize\bfseries}}
\makeatother

\usepackage{colortbl}
\usepackage{graphicx}
\usepackage{amsmath,amssymb,amsfonts,bm}
\usepackage{booktabs}
\usepackage{multirow}
\usepackage{makecell}
\usepackage{hhline}
\usepackage{arydshln}
\usepackage{float}
\usepackage{subcaption}
\usepackage[table]{xcolor}
\usepackage{algorithm}
\usepackage{algorithmic}
\usepackage{tikz}
\usetikzlibrary{positioning, shapes, arrows.meta}

\makeatletter
\def\blfootnote{\xdef\@thefnmark{}\@footnotetext}
\makeatother
\definecolor{cvprblue}{rgb}{0.21,0.49,0.74}
\usepackage[pagebackref,breaklinks,colorlinks,
            citecolor=cvprblue,
            linkcolor=cvprblue,
            urlcolor=cvprblue]{hyperref}
\renewcommand*{\backref}[1]{}
\renewcommand*{\backrefalt}[4]{%
  \ifcase #1 \or {\color{cvprblue}#2}\else {\color{cvprblue}#2}\fi}

\def\paperID{} 
\def\confName{CVPR}
\def\confYear{2026}
\definecolor{gold}{RGB}{255, 200, 0}
\definecolor{silver}{RGB}{180, 180, 180}
\definecolor{bronze}{RGB}{188, 120, 60}
\title{CANDLE: Illumination-Invariant Semantic Priors for Color \\Ambient Lighting Normalization}

\author{
Rong-Lin Jian$^{1}$ \quad Ting-Yao Chen$^{1}$ \quad Yu-Fan Lin$^{2}$ \quad Chia-Ming Lee$^{1,2}$\\
Fu-En Yang$^{3}$ \quad Yu-Chiang Frank Wang$^{3}$ \quad Chih-Chung Hsu$^{1,2}$\\[0.5em]
$^{1}$National Yang Ming Chiao Tung University \quad
$^{2}$National Cheng Kung University \quad
$^{3}$NVIDIA
}

\makeatletter
\renewcommand\paragraph{\@startsection{paragraph}{4}{\z@}%
  {4pt \@plus 1pt \@minus .2ex}%
  {-1em}%
  {\normalfont\normalsize\bfseries}}
\makeatother

\begin{document}

\maketitle

\begin{abstract}
Color ambient lighting normalization under multi-colored illumination is challenging due to severe chromatic shifts, highlight saturation, and material-dependent reflectance.
Existing geometric and low-level priors are insufficient for recovering object-intrinsic color when illumination-induced chromatic bias dominates.
We observe that DINOv3's self-supervised features remain highly consistent between colored-light inputs and ambient-lit ground truth, motivating their use as illumination-robust semantic priors.
We propose \textbf{CANDLE} (\textit{Color Ambient Normalization with DINO Layer Enhancement}), which introduces \textbf{DINO Omni-layer Guidance (D.O.G.)} to adaptively inject multi-layer DINOv3 features into successive encoder stages, and a \textbf{color-frequency refinement} design (BFACG + SFFB) to suppress decoder-side chromatic collapse and detail contamination.
Experiments on CL3AN show a +1.22 dB PSNR gain over the strongest prior method.
CANDLE achieves \textbf{3rd place} on the NTIRE 2026 ALN Color Lighting Challenge and \textbf{2nd place} in fidelity on the White Lighting track with the lowest FID, confirming strong generalization across both chromatic and luminance-dominant illumination conditions.  Code is available at \url{https://github.com/ron941/CANDLE}.
\blfootnote{This study was supported in part by the National Science and Technology Council (NSTC), Taiwan, under grants 112-2221-E-006-157-MY3, 114-2627-M-A49-003, and 114-2218-E-035-001. We thank the National Center for NCHC of NARLabs in Taiwan for providing computational and storage resources. Corresponding authors: Chih-Chung Hsu.}

\end{abstract}
\vspace{-0.5cm}

\section{Introduction}

Ambient lighting normalization (ALN) seeks to recover a uniformly illuminated image from observations corrupted by spatially varying, multi-source illumination.
While conventional shadow removal benchmarks~\cite{qu2017deshadownet,wang2018stacked} primarily address single white-light shadows, recent real-world ALN settings~\cite{ambient6k,cl3an} encompass multiple light sources, self-shadows, specular highlights, and complex material-dependent reflectance—rendering the inverse problem substantially more ill-posed.
The difficulty is further compounded under multi-colored lighting, where strong chromatic shifts, local color spill, and highlight saturation collectively preclude recovery through simple exposure correction.
Color ALN therefore demands reliable disentanglement of object-intrinsic appearance from illumination-induced chromatic contamination.

\noindent\textbf{Prior design is the key bottleneck.}
Frequency-domain methods such as IFBlend~\cite{ambient6k} exploit complementary spectral representations to improve detail recovery, while PromptNorm~\cite{promptnorm} establishes that explicit geometric priors—surface normals injected into a transformer backbone—yield substantial gains over unguided restoration.
These advances point to a shared conclusion: restoration capacity alone is insufficient; the discriminability of the guidance prior under illumination variation is equally decisive.
Yet under multi-colored illumination, geometric priors encode only local shading geometry and cannot identify what an object's intrinsic color should be once chromatic bias dominates; unguided RGB restoration models~\cite{nafnet,restormer} similarly conflate illumination-induced shifts with reflectance, producing color distortion and spatially inconsistent normalization.
The central question thus becomes: \emph{what kind of prior remains discriminative under strong chromatic illumination?}

\noindent\textbf{Key observation: DINOv3 features are illumination-invariant.}
To answer this, we measure patch-wise feature consistency between colored-light inputs and their ambient-lit ground truth across three encoders: DINOv3~\cite{dinov3}, CLIP~\cite{clip}, and supervised ResNet-50~\cite{he2016resnet}.

\begin{figure*}[t]
\centering
\includegraphics[width=0.95\linewidth]{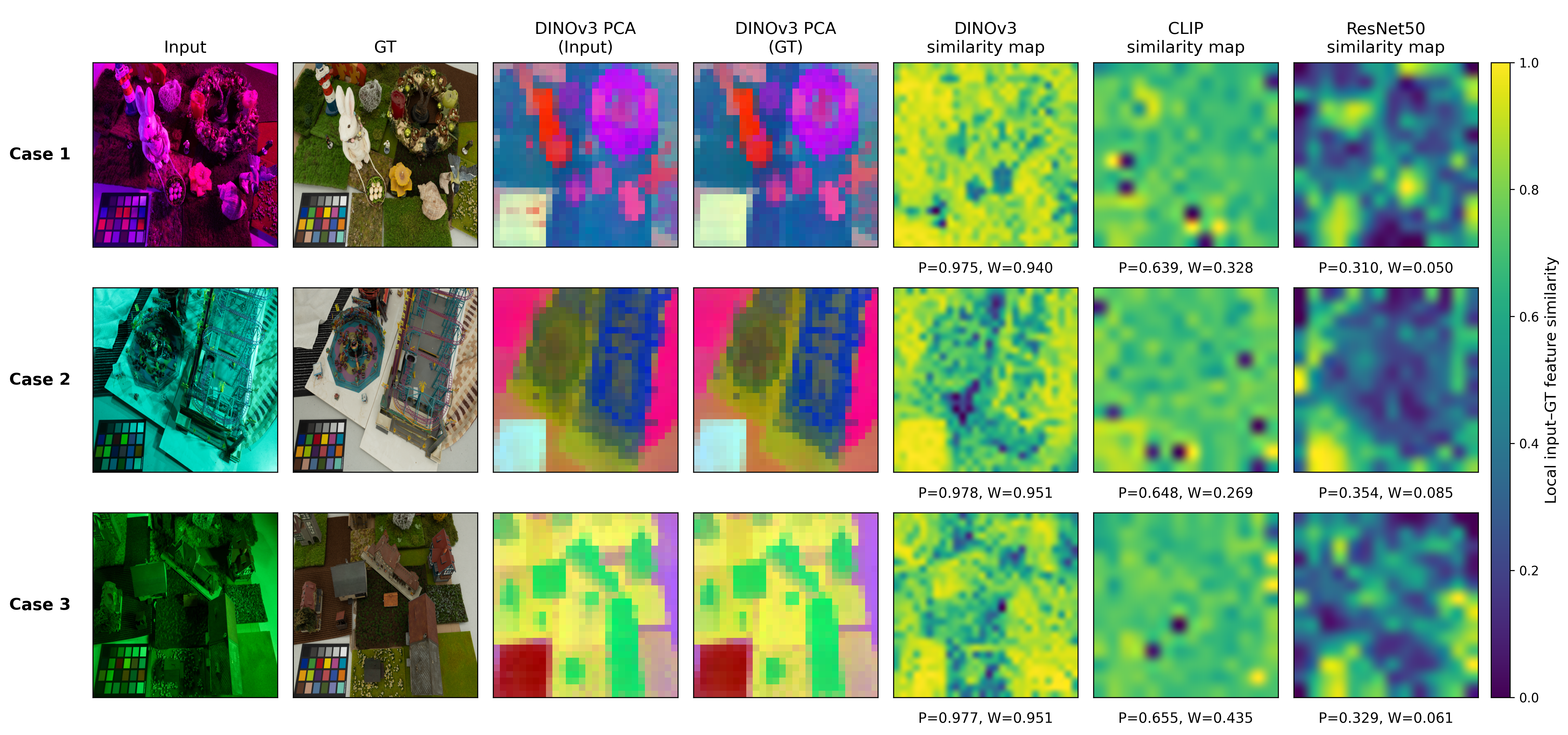}
\vspace{-4mm}
\caption{\textbf{DINOv3~\cite{dinov3} features remain highly consistent under colored illumination.}
Despite severe appearance shifts between the colored-light input and the ambient-lit GT, DINOv3 PCA visualizations remain highly consistent across domains.
Patch-wise cosine similarity maps (brighter $=$ stronger consistency) show that DINOv3 preserves substantially higher similarity both on average ($P$) and in the hardest regions ($W$), outperforming CLIP~\cite{clip} and supervised ResNet-50~\cite{he2016resnet}.}
\label{fig:fig1_representation}
\vspace{-3mm}
\end{figure*}

As illustrated in Fig.~\ref{fig:fig1_representation}, despite substantial RGB appearance divergence between domains, DINOv3 patch features remain highly consistent both globally and in locally corrupted regions, whereas CLIP and ResNet-50 exhibit pronounced feature drift under strong chromatic contamination.
\begin{figure}[t]
\centering
\includegraphics[width=\linewidth]{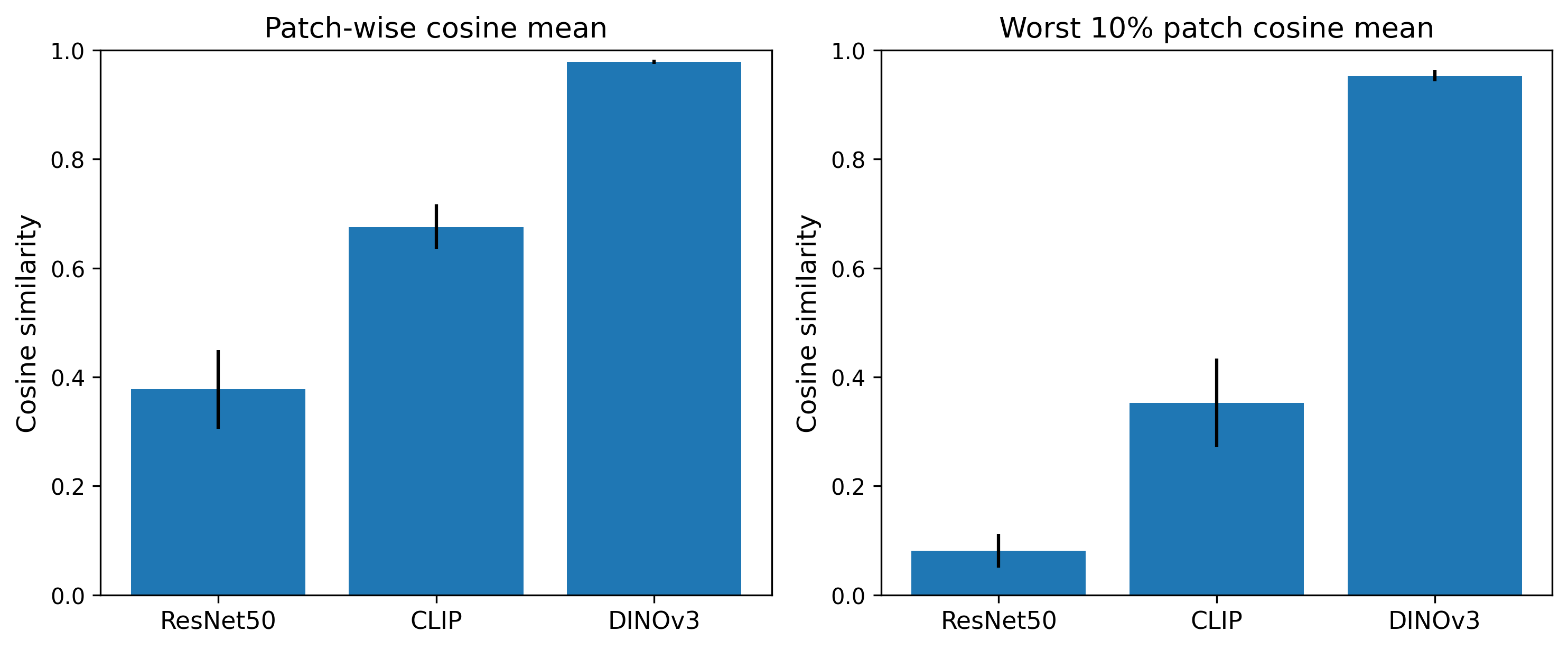}
\vspace{-4mm}
\caption{\textbf{Quantitative representation consistency over 10 input/GT pairs.}
DINOv3~\cite{dinov3} achieves the highest average patch-wise cosine similarity and the strongest robustness on the hardest 10\% of regions ($W$), consistently outperforming CLIP~\cite{clip} and ResNet-50~\cite{he2016resnet}.}
\label{fig:fig2_quantitative}
\vspace{-3mm}
\end{figure}
Fig.~\ref{fig:fig2_quantitative} quantifies this across 10 scene pairs: DINOv3 attains the highest mean patch similarity and the lowest degradation on the most severely corrupted regions.
This property stems from DINOv3's self-supervised training objective, which encourages representations invariant to appearance variation while preserving object- and material-consistent semantic structure—precisely the inductive bias required for chromatic disentanglement.

\noindent\textbf{Proposed approach: CANDLE.}
These observations motivate \textbf{CANDLE} (\textbf{C}olor \textbf{A}mbient \textbf{N}ormalization with \textbf{D}INO \textbf{L}ayer \textbf{E}nhancement), a semantic-guided restoration framework with two core contributions.
\textbf{DINO Omni-layer Guidance (D.O.G.)} replaces surface-normal injection with adaptive multi-layer DINOv3 features fused into successive encoder stages, enforcing illumination-invariant semantic consistency throughout the restoration hierarchy.
As semantic guidance constrains the encoder-side representation but does not fully govern the final reconstruction, decoder-side failure modes—chromatic collapse in saturated regions and detail contamination through illumination-corrupted skip connections—persist and must be addressed independently.
We therefore introduce a \textbf{color-frequency refinement} design comprising a bifurcated color-aware decoding module (BFACG) and a frequency-aware skip fusion block (SFFB), jointly suppressing these artifacts while preserving structural boundaries.
For the challenge submission, track-specific strategies are further incorporated, with strong white-light results confirming generalization beyond the colored-light regime. Overall, Our contributions are:
\begin{itemize}
    \item A prior-design perspective on color ALN, supported by encoder consistency analysis (Figs.~\ref{fig:fig1_representation}--\ref{fig:fig2_quantitative}), demonstrating that DINOv3~\cite{dinov3} self-supervised representations provide more illumination-robust guidance than geometric~\cite{promptnorm} or frequency-based~\cite{ambient6k} priors.
    \item \textbf{DINO Omni-layer Guidance (D.O.G.)}, replacing geometric priors with adaptive multi-layer semantic injection across encoder stages for illumination-invariant feature conditioning.
    \item A \textbf{color-frequency refinement} design (BFACG + SFFB) that decouples structural and chromatic restoration at the decoder, suppressing chromatic collapse and skip-path contamination.
    \item State-of-the-art results on CL3AN~\cite{cl3an}; \textbf{3rd-place} perceptual ranking on the NTIRE 2026 ALN Color Lighting Challenge and \textbf{2nd-place} fidelity on the White Lighting track with the lowest FID across all submissions.
\end{itemize}

\vspace{-2mm}\section{Related Works}
\label{sec:related}\vspace{-2mm}

\noindent\textbf{Image restoration.}
Modern image restoration relies on encoder–decoder architectures with multi-scale feature aggregation.
CNN-based methods (HINet~\cite{hinet}, NAFNet~\cite{nafnet}) achieve strong fidelity through simplified activations, transformer-based designs (Uformer~\cite{wang2022uformer}, Restormer~\cite{restormer}, SFNet~\cite{sfnet}) capture long-range dependencies via window or channel-wise attention, and state-space models (MambaIR~\cite{mambaIR}) offer linear-complexity alternatives.
Frequency-domain approaches~\cite{mwcnn,cheng2025wweuie,ambient6k,freqfusion} complement spatial processing by decomposing features into sub-band components, enabling targeted suppression of degradation-specific signals and improved detail recovery.
Prior-guided approaches further reduce ill-posedness: PromptIR~\cite{promptir} conditions a unified backbone with learnable prompt tokens, Retinexformer~\cite{retinexformer} decomposes illumination and reflectance following Retinex theory~\cite{land1977retinex}, and ReflexSplit~\cite{lee2026reflexsplit} demonstrates that frozen pretrained Swin features can stabilize layer disentanglement in reflection separation.
Despite these advances, all of the above methods are designed for conventional degradations and lack mechanisms to handle illumination-induced chromatic distortion under multi-colored lighting.

\noindent\textbf{Ambient lighting normalization.}
ALN~\cite{ambient6k,cl3an} extends shadow removal to recovering uniformly lit images under multiple light sources, self-shadows, highlights, and complex material-light interactions.
IFBlend~\cite{ambient6k} established a white-light ALN baseline via image-frequency fusion, RLN2~\cite{cl3an} extended this to multi-colored illumination through chromaticity–luminance disentanglement, and PromptNorm~\cite{promptnorm} demonstrated that surface normal injection into the backbone substantially improves restoration quality.
These methods, however, rely on spectral or geometric priors that remain sensitive to chromatic variation, motivating our use of illumination-invariant semantic features.

\noindent\textbf{Single image shadow removal.}
Shadow removal aims to recover true appearance beneath shadows.
Early methods relied on detection-removal pipelines with handcrafted illumination models~\cite{shor2008shadow,zhang2015shadow,cucchiara2003detecting,salamati2011removing,Le_2019_ICCV}.
Deep learning substantially advanced the field: CNNs~\cite{ronneberger2015u,qu2017deshadownet} improve local feature extraction, transformer-based methods~\cite{shadowformer,Xiao_2024_CVPR,RASM} provide better global context, and diffusion models~\cite{guo2023shadowdiffusion,mei2024latent,xu2024detailpreservinglatentdiffusionstable} achieve high perceptual quality.
More recent methods incorporate semantic and geometric priors: DeS3~\cite{des3} uses DINO-guided diffusion, OmniSR~\cite{omnisr} and DenseSR~\cite{densesr} handle direct and indirect illumination via geometric-semantic priors, and PhaSR~\cite{lee2024phasr} achieves multi-source generalization through closed-form Retinex normalization and cross-modal rectification.
These methods nonetheless remain limited to white-balanced settings and cannot resolve the global chromaticity shifts that define the color ALN problem.

\noindent\textbf{Semantic priors for restoration.}
Self-supervised vision models such as DINOv3~\cite{dinov3} encode high-level semantic structure that is inherently stable across illumination changes, offering a compelling alternative to frequency or geometric guidance.
The shadow removal methods above (DeS3, OmniSR, DenseSR) have begun adopting such features, but in the context of luminance-based degradation where geometry remains informative.
CANDLE is the first to apply multi-layer DINO features as the primary guidance signal for \emph{color} ALN, where illumination invariance is critical and geometric priors are fundamentally insufficient.

\vspace{-2mm}\section{Methodology}
\label{sec:method}\vspace{-2mm}

\begin{figure*}[t!]
\centering
\includegraphics[width=0.95\linewidth]{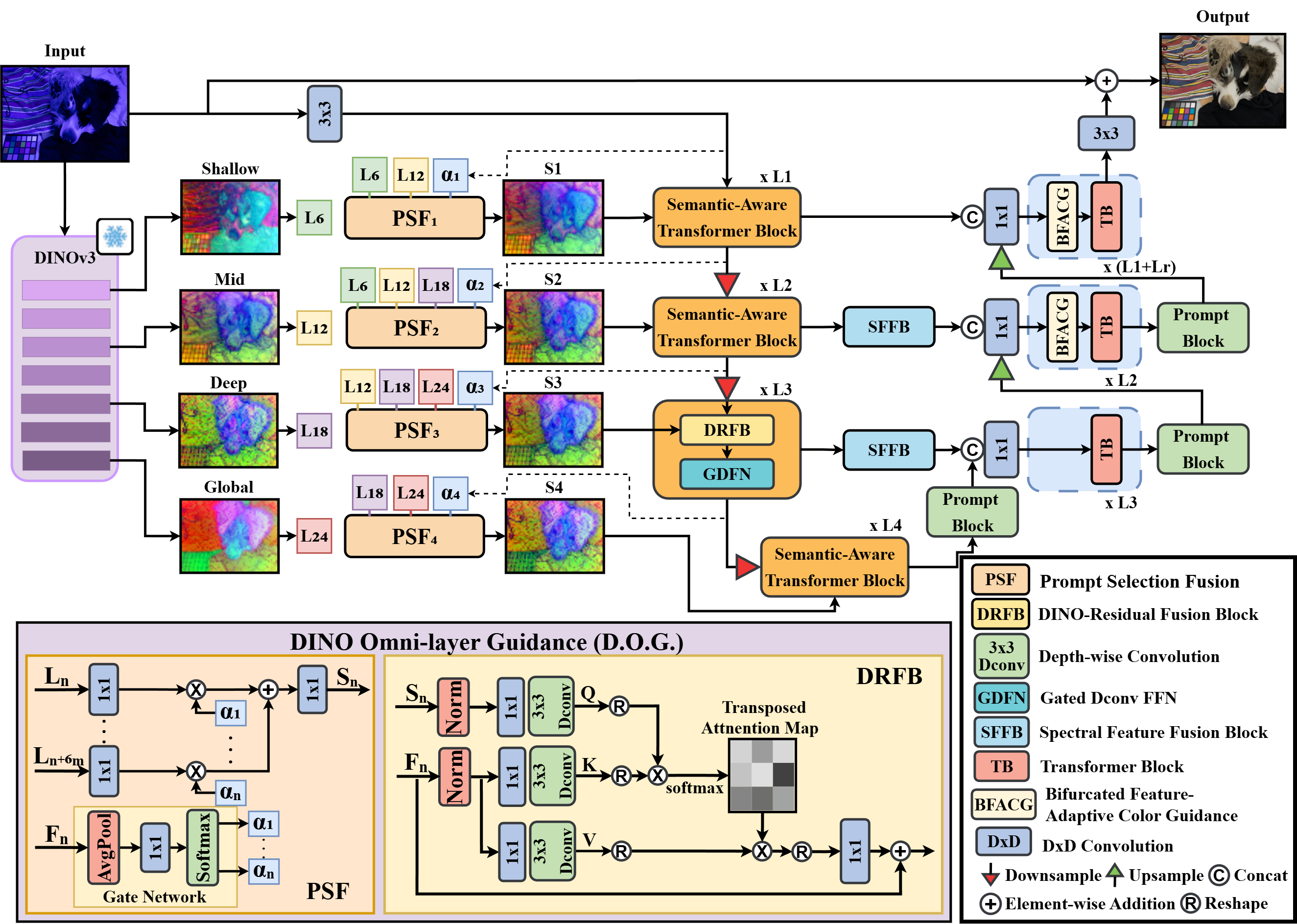}
\caption{
  \textbf{Overview of CANDLE.}
  Multi-layer features from a frozen DINOv3~\cite{dinov3} ViT-L/16 are adaptively selected and fused via PSF and injected into successive encoder stages through DRFB (D.O.G., Sec.~\ref{sec:dog}), replacing surface-normal guidance with illumination-robust semantic priors.
  At the decoder, BFACG decouples structural and chromatic restoration under edge-aware modulation, while SFFB suppresses illumination-corrupted skip features via Haar wavelet gating (Sec.~\ref{sec:cfr}).
}
\label{fig:pipeline}
\vspace{-3mm}
\end{figure*}

\paragraph{Overview.} Color ambient lighting normalization under multi-colored illumination requires recovering object-intrinsic appearance once illumination-induced chromatic bias is removed—a fundamentally different challenge from geometric shadow removal.
Existing methods rely on geometric priors~\cite{promptnorm} or chromaticity-oriented representations~\cite{ambient6k,cl3an}, both of which fail when strong chromatic shifts decouple apparent color from material identity.
As illustrated in Fig.~\ref{fig:pipeline}, CANDLE addresses this through two components: \textbf{DINO Omni-layer Guidance (D.O.G.)}, which fuses multi-layer DINOv3 features into successive encoder stages as illumination-robust semantic priors, and \textbf{color-frequency refinement}, which suppresses chromatic collapse and detail contamination at the decoder.
The network predicts the ambient-normalized output via a global residual, $\hat{\mathbf{Y}} = \mathbf{I} + f_{\theta}(\mathbf{I})$, where $f_{\theta}$ denotes the full restoration network.

\vspace{-2mm}\subsection{DINO Omni-layer Guidance}
\label{sec:dog}\vspace{-2mm}

Motivated by the illumination invariance demonstrated in Figs.~\ref{fig:fig1_representation}--\ref{fig:fig2_quantitative}, D.O.G.\ extracts features from a frozen DINOv3~\cite{dinov3} ViT-L/16 at layers $\mathcal{L} = \{6, 12, 18, 24\}$.
Integrating across this hierarchy provides complementary cues at different granularities: shallow layers preserve spatially precise material boundaries while deep layers encode illumination-invariant object identity.
D.O.G.\ achieves this through adaptive multi-layer selection via Prompt Selection Fusion (PSF), followed by semantic injection via the DINO-Residual Fusion Block (DRFB).

\paragraph{Prompt Selection Fusion.} 
Prior guidance methods~\cite{promptnorm,promptir,densesr} inject a single fixed-layer feature into the restoration backbone—surface normals in PromptNorm~\cite{promptnorm} or task-specific geometric-semantic features in DenseSR~\cite{densesr}—ignoring that different encoder stages operate at different spatial scales and semantic granularities.
A shallow stage recovering fine-grained texture benefits from low-level DINOv3~\cite{dinov3} features, whereas a deeper stage resolving global color consistency requires higher-level abstraction.
PSF addresses this by adaptively selecting and fusing the most relevant DINO layers for each encoder stage.

Concretely, for encoder stage $s$, each DINO layer feature $F^{(l)}$ is first spatially resized to match the stage's resolution and projected to the backbone channel dimension via $\phi(\cdot)$.
Gating weights $\alpha_{s,l}$ are then predicted from the current backbone feature $F_{\mathrm{enc}}^{(s)}$ via global average pooling followed by a small MLP, producing a stage-conditioned soft selection over layers.
The fused semantic feature is:
\begin{equation}
    S_s = \sum_{l \in \mathcal{L}_s} \alpha_{s,l} \cdot \phi(F^{(l)}), \quad \sum_l \alpha_{s,l} = 1,
\end{equation}
where the normalized weights allow each stage to dynamically emphasize the semantic level most relevant to its restoration function.
The resulting $S_s$ is then passed to DRFB for cross-attention injection into the backbone (described below).

\paragraph{DINO-Residual Fusion Block.}
With $S_s$ providing stage-appropriate semantic content, DRFB injects it into the backbone through query-based cross-attention~\cite{promptnorm}, which is more effective than naive concatenation or additive fusion for conditioning restoration on external guidance~\cite{promptir}. DRFB uses backbone features as queries and the semantic prior $S_s$ as both keys and values:
\begin{align}
    Q &= W_Q F_{\mathrm{enc}}, \quad K = W_K S_s, \quad V = W_V S_s, \\
    F_{\mathrm{enc}}' &= F_{\mathrm{enc}} + \gamma \cdot \mathrm{softmax}\!\left(\frac{QK^\top}{\sqrt{d}}\right)V,
\end{align}
where $W_Q, W_K, W_V$ are learned projections and $\gamma$ is a zero-initialized scale factor~\cite{bachlechner2021rezero}.
This formulation allows each spatial location in the backbone to selectively retrieve the most semantically consistent region of the DINO prior, while the zero-initialized $\gamma$ preserves the pretrained backbone dynamics at initialization and allows semantic conditioning to be introduced gradually during training.

Compared with PromptIR~\cite{promptir}, which uses learnable task-agnostic prompt tokens, and PromptNorm~\cite{promptnorm} and DenseSR~\cite{densesr}, which rely on geometric or supervised priors sensitive to illumination variation, D.O.G.\ derives its guidance from self-supervised DINOv3~\cite{dinov3} features that encode object-consistent semantics stably across chromatic conditions—providing a more discriminative reference for intrinsic color recovery under multi-colored illumination.

\vspace{-2mm}\subsection{Color-Frequency Refinement}
\label{sec:cfr}\vspace{-2mm}
Semantic guidance constrains global appearance consistency but does not prevent two localized decoder-side failure modes: chromatic collapse in highlight-heavy regions, and detail contamination propagated through skip connections carrying illumination-corrupted encoder features.
We address both through a pair of complementary modules: Bifurcated Feature-Adaptive Color Guidance (BFACG) for decoder-side disentanglement, and Spectral Feature Fusion Block (SFFB) for skip-connection filtering.

\paragraph{Bifurcated Feature-Adaptive Color Guidance.}
Existing restoration backbones~\cite{nafnet,restormer} and ALN methods~\cite{ambient6k,promptnorm} process structural and chromatic information jointly in a shared decoder stream, conflating geometry-driven detail recovery with illumination-driven color correction.
BFACG explicitly separates these objectives—motivated by disentangled restoration~\cite{cheng2025wweuie,lee2026reflexsplit}—by routing each intermediate decoder feature $F$ through two parallel convolutional branches:
\begin{equation}
    F_{\mathrm{str}} = \mathrm{Conv}_{\mathrm{str}}(F), \qquad F_{\mathrm{chr}} = \mathrm{Conv}_{\mathrm{chr}}(F),
\end{equation}
where $F_{\mathrm{str}}$ focuses on geometry-consistent detail recovery and $F_{\mathrm{chr}}$ on illumination-invariant color correction.
To prevent color correction from bleeding into structural boundaries, an edge-aware guidance map $G_{\mathrm{edge}} \in [0,1]^{H\times W}$—derived from the gradient magnitude of $F$—modulates the branch fusion:
\begin{equation}
    F' = G_{\mathrm{edge}} \odot F_{\mathrm{str}} + (1 - G_{\mathrm{edge}}) \odot F_{\mathrm{chr}},
\end{equation}
where $\odot$ denotes element-wise multiplication.
High-gradient (boundary) regions are steered toward the structural branch, while low-gradient (flat) regions are steered toward chromatic correction.

\paragraph{Spectral Feature Fusion Block.}
Prior frequency-domain methods~\cite{mwcnn,freqfusion,cheng2025wweuie} apply spectral processing to intermediate or upsampled features, but do not specifically address the illumination bias reintroduced by skip connections.
In standard encoder–decoder architectures~\cite{unet}, skip connections relay encoder features directly to the decoder, bypassing D.O.G.'s semantic correction.
SFFB intercepts each skip feature $F_{\mathrm{skip}}$ via Haar wavelet decomposition~\cite{haar1910theorie}, separating it into a low-frequency subband $F_{\mathrm{LL}}$ encoding global color bias and high-frequency subbands $\{F_{\mathrm{LH}}, F_{\mathrm{HL}}, F_{\mathrm{HH}}\}$ encoding structural detail:
\begin{equation}
    \{F_{\mathrm{LL}},\, F_{\mathrm{LH}},\, F_{\mathrm{HL}},\, F_{\mathrm{HH}}\} = \mathcal{W}(F_{\mathrm{skip}}).
\end{equation}
A learned gate selectively suppresses illumination-sensitive low-frequency components:
\begin{equation}
    F_{\mathrm{LL}}' = \sigma(W_g \ast F_{\mathrm{LL}}) \odot F_{\mathrm{LL}},
\end{equation}
where $\sigma(\cdot)$ is a sigmoid activation and $W_g$ is a learned convolutional filter.
The high-frequency subbands are retained, and the filtered feature is reconstructed via inverse transform:
\begin{equation}
    \tilde{F}_{\mathrm{skip}} = \mathcal{W}^{-1}(F_{\mathrm{LL}}',\, F_{\mathrm{LH}},\, F_{\mathrm{HL}},\, F_{\mathrm{HH}}).
\end{equation}
Together, BFACG and SFFB stabilize decoder-side reconstruction: the former disentangles structural and chromatic restoration within the decoder stream, while the latter intercepts illumination leakage at the skip-connection level.

\vspace{-2mm}\subsection{Training Objective and Strategy}
\label{sec:training}
\vspace{-2mm}
\noindent\textbf{Loss Function.}
We supervise reconstruction using an L1 loss combined with SSIM loss~\cite{wang2004image}    for structural consistency:
\begin{equation}
    \mathcal{L} = \|\hat{\mathbf{Y}} - \mathbf{Y}_{\text{GT}}\|_1
    + 0.7 \left(1 - \text{SSIM}(\hat{\mathbf{Y}}, \mathbf{Y}_{\text{GT}})\right).
\end{equation}

\noindent\textbf{Competition-Oriented Strategy.}
Beyond the core architecture, we adopt a set of \textit{track-specific optimization strategies} for the final challenge submission.
These components are designed for practical performance improvement and are not part of the core model design.

\paragraph{Color-light training pipeline.}
For the color-light setting, we employ a three-stage training pipeline to progressively refine restoration quality.
\textit{Stage 1} trains the full CANDLE architecture under a progressive patch-size curriculum, encouraging both local detail recovery and large-scale illumination consistency.
\textit{Stage 2} freezes the Stage-1 backbone and trains a lightweight NAFNet-based~\cite{nafnet} refiner $g_{\phi}$ to predict a residual correction:
\begin{equation}
    \hat{\mathbf{Y}}_{\text{ref}} =
    \mathrm{clip}\!\left(\hat{\mathbf{Y}}_{\text{S1}} + g_{\phi}(\hat{\mathbf{Y}}_{\text{S1}}),\; 0, 1\right).
\end{equation}
\textit{Stage 3} jointly fine-tunes the backbone and refiner end-to-end, allowing them to co-adapt toward the final perceptual objective.

In addition, we incorporate several practical strategies, including \textit{retrieval-based finetuning}, the use of \textit{additional synthetic color-light data} generated from Ambient6K, and joint optimization with auxiliary refinement networks, to further improve performance under the competition setting.

\paragraph{White-light setting.}
In contrast to the color-light track, the white-light setting benefits less from training-stage expansion and more from model complementarity.
We therefore adopt an ensemble strategy combining multiple models with different inductive biases, which leads to improved robustness and performance under standard ALN conditions.


\vspace{-2mm}\section{Experiment Results}
\vspace{-2mm}
\label{sec:setup}

\noindent\textbf{Datasets.}
CL3AN~\cite{cl3an} is the primary benchmark used in this work and also serves as the basis of the NTIRE 2026 color-light setting, consisting of high-resolution scene triplets captured under multiple RGB directional lights paired with ambient-lit references.
It features strong chromatic shifts, highlight saturation, color spill, and material-dependent reflectance distortions, making it substantially more challenging than conventional shadow removal or white-light ALN benchmarks.
We follow the official split with 3,667 training, 437 validation, and 431 test images. The dataset is captured at approximately 24MP resolution (6000 $\times$ 4000).

Ambient6K~\cite{ambient6k} contains 6,000 images of scenes under up to three white-aligned LED directional lights, with a native resolution of $2560 \times 1440$, and is relevant to the standard white-light ALN setting considered in the NTIRE 2026 white-light track.

\noindent\textbf{Metrics.}
We report PSNR and SSIM~\cite{wang2004image} for restoration fidelity, and LPIPS~\cite{zhang2018perceptual} for perceptual quality.
For the challenge evaluation, FID~\cite{heusel2017gans} is additionally reported.

\noindent\textbf{Complexity.}
The computational complexity, reported as GMACs, is measured on $128 \times 128$ input patches using \texttt{ptflops} for consistency. Unless otherwise specified, all compared methods are evaluated under the same input size and counting protocol for fair efficiency comparison. These values are intended to reflect relative computational cost and do not represent full-resolution inference complexity on the original benchmark images.

\noindent\textbf{Implementation Details.}
CANDLE is built upon the PromptNorm~\cite{promptnorm} backbone, replacing surface-normal guidance with D.O.G. and adding color-frequency refinement at the decoder side.
We use frozen DINOv3 ViT-L/16 features extracted from layers $\{6, 12, 18, 24\}$ as semantic priors.
All images are first resized to $1024 \times 768$, and training is performed using random $512 \times 512$ crops. For the full competition system, we adopt a progressive resizing strategy, where the training resolution is gradually increased from $256$ to $512$ and finally to $768$.
We use Adam with stage-wise learning rates ($1\times10^{-4} \rightarrow 5\times10^{-5} \rightarrow 2\times10^{-5}$) and cosine annealing in later stages.
All experiments use random horizontal and vertical flipping, as well as random $90^\circ$ rotations for data augmentation.
For the controlled ablation setting (Sec.~\ref{sec:ablation}), all methods are trained under a unified single-stage protocol without extra data or multi-stage refinement, ensuring fair architectural comparison.
For the full competition system used in the color-light setting (Sec.~\ref{sec:competition}), we additionally incorporate extra synthetic color-light data generated from Ambient6K and the competition-oriented strategies described in Sec.~\ref{sec:training}.
All experiments are conducted on NVIDIA A100 GPUs.
\vspace{-2mm}\subsection{Main Results on CL3AN}
\label{sec:main_cl3an}
\vspace{-2mm}
\begin{figure*}[t!]
\centering
\includegraphics[width=1\linewidth]{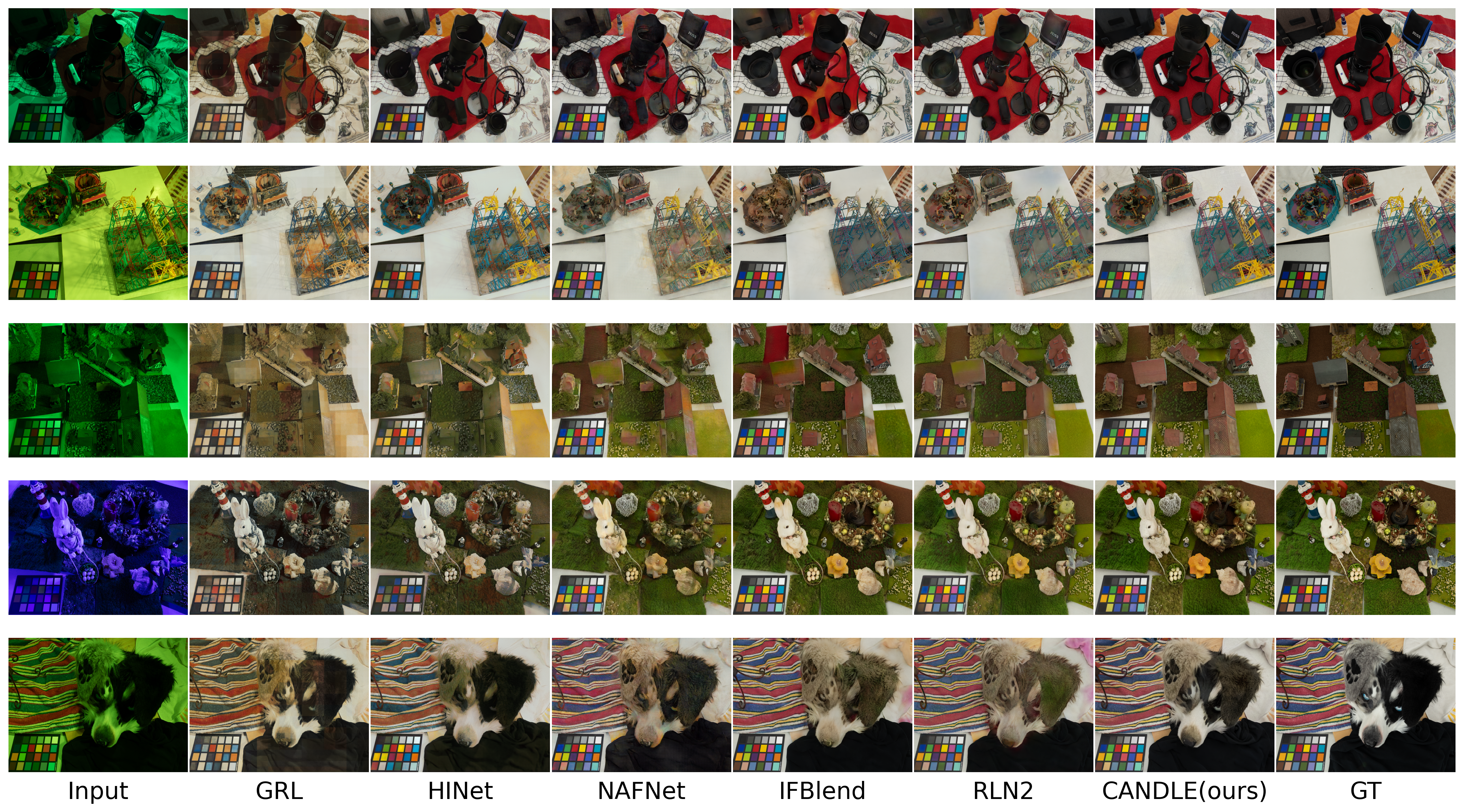}
\caption{\textbf{Qualitative comparison with state-of-the-art methods on CL3AN~\cite{cl3an}.}
General restoration methods~\cite{nafnet,restormer} exhibit washed-out colors and incomplete correction under strong chromatic shifts.
ALN-specific methods~\cite{ambient6k,promptnorm} partially recover global color balance but remain inconsistent across materials, as frequency and geometric priors are insufficient when chromaticity shifts dominate.
CANDLE recovers object-intrinsic color more faithfully with sharper boundaries, enabled by D.O.G.\ and color-frequency refinement.
}
\label{fig:comparison}
\vspace{-3mm}
\end{figure*}

Table~\ref{tab:cl3an_main} compares CANDLE against representative restoration and ALN methods on CL3AN.

\begin{table}[t]
\centering
\caption{\textbf{Quantitative comparison on the CL3AN benchmark.}
\colorbox{gold!40}{\strut 1st}~\colorbox{silver!50}{\strut 2nd}~\colorbox{bronze!30}{\strut 3rd} indicate top-three methods by PSNR.}
\label{tab:cl3an_main}
\vspace{0.2em}
\resizebox{0.48\textwidth}{!}{
\begin{tabular}{lccccc}
\toprule
Method & Venue & MACs(G.) & PSNR $\uparrow$ & SSIM $\uparrow$ & LPIPS $\downarrow$ \\
\midrule
NAFNet~\cite{nafnet}               & ECCV 2022  & 4.04 & 18.547 & 0.6121 & 0.3804 \\
SFNet~\cite{sfnet}                 & ICLR 2023  & 30.59 & 16.778 & 0.5489 & 0.6183 \\
Uformer~\cite{wang2022uformer}     & CVPR 2022  & 20.89 & 17.785 & 0.6815 & 0.3437 \\
Restormer~\cite{restormer}         & CVPR 2022  & 35.31 & 18.244 & 0.6570 & 0.3541 \\
HINet~\cite{hinet}                 & ICCV 2021  & 39.46 & 17.728 & 0.6035 & 0.3887 \\
\rowcolor{bronze!30}
IFBlend~\cite{ambient6k}           & ECCV 2024  & 24.65 & 19.418 & 0.7581 & 0.2505 \\
MambaIR~\cite{mambaIR}             & ECCV 2024  & 219.38 & 16.294 & 0.6472 & 0.4098 \\
Retinexformer~\cite{retinexformer} & ICCV 2023  & 3.96 & 18.818 & 0.6943 & 0.3167 \\
\rowcolor{silver!40}
RLN2-Lf~\cite{cl3an}              & ICCV 2025  & 21.93 & 19.849 & 0.7436 & 0.2569 \\
PromptNorm~\cite{promptnorm}       & CVPRW 2025 & 20.81 & 19.216 & 0.7485 & 0.2590 \\
\midrule
\rowcolor{gold!40}
CANDLE (ours)                      & ---        & 23.52 & \textbf{21.066} & \textbf{0.7788} & \textbf{0.2325} \\
\bottomrule
\end{tabular}}
\vspace{-0.4em}
\end{table}

General restoration methods~\cite{nafnet,restormer,wang2022uformer} struggle on CL3AN, as severe chromatic shifts and material-dependent color spill lie beyond the scope of standard exposure correction.
Among ALN-specific baselines, IFBlend~\cite{ambient6k} achieves competitive results via frequency-domain fusion but cannot distinguish illumination-induced color variation from true material identity, while PromptNorm~\cite{promptnorm} is limited by geometric priors that are ineffective when chromaticity shifts dominate shading geometry (19.216 dB).
RLN2~\cite{cl3an} reaches the strongest prior result at 19.849 dB through explicit chromaticity–luminance disentanglement.
CANDLE surpasses all baselines with 21.066 dB PSNR—a +1.22 dB gain over RLN2 and +1.85 dB over PromptNorm—by replacing geometric priors with illumination-robust semantic guidance and suppressing decoder-side chromatic collapse via color-frequency refinement.
Qualitatively (Fig.~\ref{fig:comparison}), prior methods exhibit incomplete correction in strongly illuminated regions and inconsistent restoration across materials; CANDLE better preserves object-intrinsic color with sharper structural boundaries.

\vspace{-2mm}\subsection{Architectural Ablation}
\label{sec:ablation}
\vspace{-2mm}
All experiments below use a \textbf{unified controlled setting}—identical resize, crop, augmentation, and optimization—without extra data, progressive curriculum, or Stage-2/3 extensions, so that each comparison isolates a single architectural variable.
All results are on the CL3AN test split.

\noindent\textbf{Guidance Representation.}
Table~\ref{tab:abl_guidance_rep} compares three guidance configurations.
Replacing the unguided baseline with geometric normal guidance (PromptNorm~\cite{promptnorm}) yields +1.77 dB, confirming the value of explicit priors.
Replacing normals with our DINO Omni-layer Guidance (D.O.G.) yields a further +1.85 dB, demonstrating that under multi-colored illumination, illumination-robust semantic priors are substantially more effective than local shading geometry for recovering intrinsic color.

\begin{table}[t]
\centering
\caption{\textbf{Ablation on guidance representation.}
Semantic guidance outperforms geometric normal guidance for color ALN.}
\label{tab:abl_guidance_rep}
\vspace{0.2em}
\resizebox{0.42\textwidth}{!}{
\begin{tabular}{lccc}
\toprule
Configuration & PSNR $\uparrow$ & SSIM $\uparrow$ & LPIPS $\downarrow$ \\
\midrule
w/o guidance                      & 17.449 & 0.7131 & 0.3019 \\
Normal Guidance~\cite{promptnorm} & 19.216 & 0.7485 & 0.2590 \\
\rowcolor{gray!15}
CANDLE (D.O.G., ours)             & \textbf{21.066} & \textbf{0.7788} & \textbf{0.2325} \\
\bottomrule
\end{tabular}}
\vspace{-0.4em}
\end{table}

\noindent\textbf{Guidance Integration.}
Table~\ref{tab:abl_guidance_int} compares four strategies for injecting DINO features into the backbone, all including BFACG and SFFB for a fair comparison.
Naive concatenation and feature/value fusion provide limited control over how semantic priors steer restoration and plateau near 20.5--20.7 dB.
Query-based integration, which directly injects DINO features as queries, fails to outperform these strategies, indicating that naive query injection is insufficient for effectively leveraging semantic priors, likely due to the lack of structured and multi-layer guidance.
Our full D.O.G. further adds adaptive multi-layer selection and progressive residual fusion, yielding an additional gain and the best overall result, confirming that \emph{both} components are necessary.

\begin{figure}[t!]
\centering
\includegraphics[width=\linewidth]{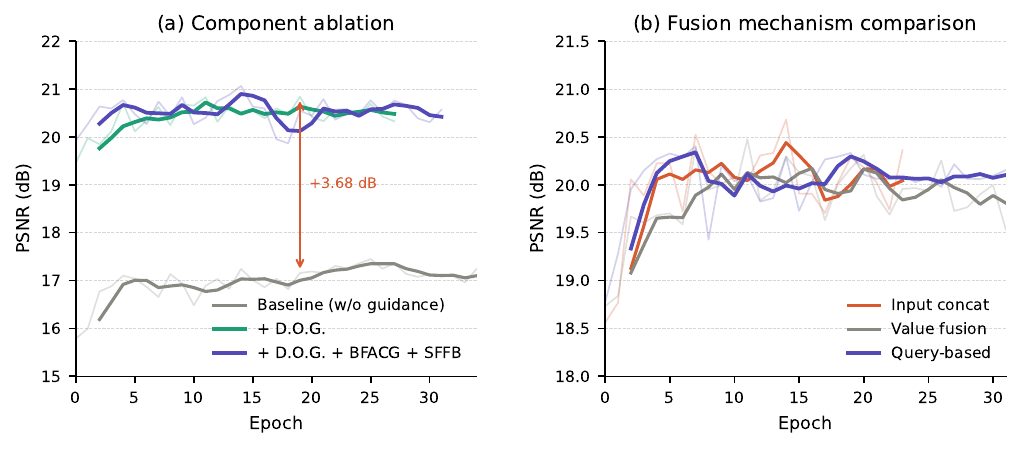}
\caption{\textbf{Training dynamics of ablation variants.}
(a)~D.O.G.\ sustains a peak gain of {+3.68}\,dB over the unguided baseline, 
with the gap maintained across all training epochs, confirming a structural 
rather than incidental improvement.
(b)~Query-based fusion converges more stably than input-concat or value-fusion variants in later epochs.}
\label{fig:ablation_curves}
\end{figure}

Fig.~\ref{fig:ablation_curves}(b) shows that query-based variants also converge more stably than concat or value-fusion in later epochs.

\begin{table}[t]
\centering
\caption{\textbf{Ablation on guidance integration strategies.}
All variants include BFACG and SFFB.
D.O.G.\ with adaptive multi-layer fusion achieves the best performance.}
\label{tab:abl_guidance_int}
\vspace{-2mm}
\resizebox{0.42\textwidth}{!}{
\begin{tabular}{lccc}
\toprule
Integration & PSNR $\uparrow$ & SSIM $\uparrow$ & LPIPS $\downarrow$ \\
\midrule
Input concat          & 20.682 & 0.7649 & 0.2380 \\
Feature/value fusion  & 20.475 & 0.7613 & 0.2408 \\
Query-based           & 20.403 & 0.7628 & 0.2418 \\
\rowcolor{gray!15}
CANDLE (D.O.G., ours) & \textbf{21.066} & \textbf{0.7788} & \textbf{0.2325} \\
\bottomrule
\end{tabular}}
\vspace{-2mm}
\end{table}

\noindent\textbf{Architecture Build-up.}
Table~\ref{tab:abl_buildop} incrementally adds each component.
D.O.G.\ alone accounts for +3.39 dB over the unguided baseline, establishing semantic guidance as the dominant contribution.
Color-frequency refinement (BFACG + SFFB) adds a further +0.23 dB PSNR alongside perceptual improvement (LPIPS $0.2334 \to 0.2325$), confirming that decoder-side reconstruction quality is complementary to encoder-side priors.
Fig.~\ref{fig:ablation_curves}(a) shows this gap is present from the first epoch and grows consistently, ruling out a late-stage or coincidental effect.

\begin{table}[t]
\centering
\caption{\textbf{Architecture build-up.} Each component contributes consistently to final performance.}
\label{tab:abl_buildop}
\vspace{0.2em}
\resizebox{0.48\textwidth}{!}{
\begin{tabular}{lccc}
\toprule
Configuration & PSNR $\uparrow$ & SSIM $\uparrow$ & LPIPS $\downarrow$ \\
\midrule
Baseline (w/o guidance)              & 17.449 & 0.7131 & 0.3019 \\
$+$ D.O.G.                           & 20.840 & 0.7705 & 0.2334 \\
\rowcolor{gray!15}
$+$ D.O.G.$+$BFACG$+$SFFB (CANDLE)  & \textbf{21.066} & \textbf{0.7788} & \textbf{0.2325} \\
\bottomrule
\end{tabular}}
\vspace{-0.4em}
\end{table}

\vspace{-2mm}\subsection{NTIRE 2026 Challenge Results}
\label{sec:competition}
\vspace{-2mm}
We evaluate the full competition system on the NTIRE 2026 ALN Challenge under the official perceptual ranking protocol~\cite{ntire2026ambient}, covering both color and white lighting tracks.

\paragraph{Color Lighting Track.}
Table~\ref{tab:color_track} shows the effect of incrementally stacking complementary strategies on CANDLE.
Retrieval-based finetuning via DINO-CLIP similarity~\cite{dinov3,clip} aligns the training distribution with test-time semantics (+0.30 dB, $-$0.008 LPIPS), additional synthetic color-light data from Ambient6K ($\sim$300 images) expands the effective training set (+0.30 dB), and joint training with a NAFNet~\cite{nafnet} refiner further improves local perceptual quality, yielding the best result (LPIPS 0.1964).

\begin{table}[t]
\centering
\caption{\textbf{Effect of auxiliary strategies on the NTIRE 2026 ALN Color Lighting track.}}
\label{tab:color_track}
\vspace{0.2em}
\resizebox{0.48\textwidth}{!}{
\begin{tabular}{lccc}
\toprule
Configuration & PSNR $\uparrow$ & SSIM $\uparrow$ & LPIPS $\downarrow$ \\
\midrule
CANDLE                                                & 20.779 & 0.7250 & 0.2159 \\
$+$ DINO-CLIP Retrieval~\cite{dinov3,clip}            & 21.079 & 0.7291 & 0.2079 \\
$+$ Extra synthetic data                        & 21.376 & 0.7365 & 0.2021 \\
$+$ NAFNet~\cite{nafnet} Joint Training               & \textbf{21.432} & \textbf{0.7431} & \textbf{0.1964} \\
\bottomrule
\end{tabular}}
\vspace{-0.4em}
\end{table}

\paragraph{White Lighting Track.}
Unlike the color track, the white lighting setting benefits more from model complementarity than training expansion.
CANDLE already outperforms DenseSR~\cite{densesr} by +0.59 dB PSNR, where degradation is dominated by luminance rather than chromaticity shifts.
Ensembling the two models yields a substantial jump in both fidelity (+0.33 dB) and perceptual quality (LPIPS $0.1072 \to 0.0892$), suggesting they capture complementary aspects of structural and chromatic restoration.

\begin{table}[t]
\centering
\caption{\textbf{Effect of ensemble strategies on the NTIRE 2026 ALN White Lighting track.}}
\label{tab:white_track}
\vspace{0.2em}
\resizebox{0.48\textwidth}{!}{
\begin{tabular}{lccc}
\toprule
Configuration & PSNR $\uparrow$ & SSIM $\uparrow$ & LPIPS $\downarrow$ \\
\midrule
DenseSR~\cite{densesr}                                & 23.785 & 0.8466 & 0.1007 \\
CANDLE                                                & 24.370 & 0.8516 & 0.1078 \\
$+$ DINO-CLIP Retrieval~\cite{dinov3,clip}            & 24.542 & 0.8533 & 0.1072 \\
$+$ DenseSR~\cite{densesr} Ensemble                  & \textbf{24.875} & \textbf{0.8639} & \textbf{0.0892} \\
\bottomrule
\end{tabular}}
\vspace{-0.4em}
\end{table}

\paragraph{Overall Challenge Performance.}
Tables~\ref{tab:ntire_color} and~\ref{tab:ntire_white} report the full leaderboard.
CANDLE (ACVLAB) ranks 3rd in perceptual ranking on the color track with the 2nd-best LPIPS (0.1964), and 2nd in fidelity and 3rd in perceptual ranking on the white track with the lowest FID (49.838) among all submissions.
The gap to the top perceptual rank is modest ($\Delta$LPIPS $<$0.01 on color, $<$0.001 on white), while CANDLE maintains a clear margin over the majority of competing methods, confirming consistent generalization across both chromatic and luminance-dominant illumination conditions.

\begin{table}[t]
\centering
\caption{\textbf{Ranking on NTIRE 2026 ALN Color Lighting Challenge.}
Results follow the official NTIRE 2026 challenge report~\cite{ntire2026ambient}.}
\label{tab:ntire_color}
\vspace{0.2em}
\resizebox{0.48\textwidth}{!}{
\begin{tabular}{lcccccc}
\toprule
Method & PSNR $\uparrow$ & SSIM $\uparrow$ & LPIPS $\downarrow$ & FID $\downarrow$ & Fidelity & Perceptual \\
\midrule
MiPorAlgo             & 21.305 & 0.6294 & 0.2478 & 92.900  & 4  & 1 \\
SNU-ISPL-B            & 22.038 & 0.7528 & 0.1861 & 82.013  & 1  & 2 \\
\rowcolor{gray!15}
ACVLAB (Ours)         & 21.432 & 0.7431 & 0.1964 & 88.200  & 3  & 3 \\
SNU-ISPL-A            & 21.900 & 0.7436 & 0.2038 & 86.654  & 2  & 4 \\
OUT\_OF\_MEMORY       & 20.175 & 0.6637 & 0.3338 & 124.439 & 5  & 5 \\
Amber                 & 17.863 & 0.6812 & 0.2835 & 149.271 & 6  & 6 \\
IllumiBlend           & 17.696 & 0.6137 & 0.3510 & 184.043 & 7  & 7 \\
MMAINAF               & 16.863 & 0.5785 & 0.3887 & 187.681 & 8  & 8 \\
miketjc               & 16.717 & 0.5607 & 0.4985 & 183.655 & 9  & 9 \\
KLETech-CEVI          & 11.935 & 0.4475 & 0.5688 & 222.755 & 10 & 10 \\
\bottomrule
\end{tabular}}
\vspace{-0.4em}
\end{table}

\begin{table}[t]
\centering
\caption{\textbf{Ranking on NTIRE 2026 ALN White Lighting Challenge.}
Results follow the official NTIRE 2026 challenge report~\cite{ntire2026ambient}.}
\label{tab:ntire_white}
\vspace{0.2em}
\resizebox{0.48\textwidth}{!}{
\begin{tabular}{lcccccc}
\toprule
Method & PSNR $\uparrow$ & SSIM $\uparrow$ & LPIPS $\downarrow$ & FID $\downarrow$ & Fidelity & Perceptual \\
\midrule
SNU-ISPL-A            & 24.275 & 0.8536 & 0.0986 & 54.169  & 6  & 1 \\
MILab\_ALN            & 26.680 & 0.8728 & 0.0832 & 50.218  & 1  & 2 \\
\rowcolor{gray!15}
ACVLAB (Ours)         & 24.875 & 0.8639 & \underline{0.0892} & \textbf{49.838} & 2  & 3 \\
MiPorAlgo             & 23.141 & 0.7631 & 0.1687 & 59.441  & 11 & 4 \\
OUT\_OF\_MEMORY       & 25.416 & 0.8604 & 0.0938 & 54.496  & 3  & 5 \\
SNU-ISPL-B            & 24.300 & 0.8561 & 0.1006 & 51.883  & 5  & 6 \\
UniBlendNet           & 24.742 & 0.8505 & 0.1052 & 53.563  & 7  & 7 \\
USTC-Seer             & 25.081 & 0.8579 & 0.0963 & 56.705  & 4  & 8 \\
GeoNorm               & 20.562 & 0.7869 & 0.1431 & 86.479  & 8  & 9 \\
KLETech-CEVI          & 20.547 & 0.7948 & 0.1494 & 87.033  & 9  & 10 \\
Light Tamers          & 20.415 & 0.7909 & 0.1487 & 90.221  & 10 & 11 \\
IllumiBlend           & 19.885 & 0.7802 & 0.1537 & 95.922  & 12 & 12 \\
NTR                   & 19.395 & 0.7649 & 0.1789 & 104.246 & 13 & 13 \\
IVPPROJECT            & 18.173 & 0.7189 & 0.2096 & 114.230 & 14 & 14 \\
\bottomrule
\end{tabular}}
\vspace{-0.4em}
\end{table}

\vspace{-2mm}
\section{Conclusion}
\label{sec:conclusion}
\vspace{-2mm}

We presented CANDLE, a framework for color ambient lighting normalization guided by illumination-invariant semantic priors.
DINO Omni-layer Guidance (D.O.G.) replaces geometric surface normal injection with adaptive multi-layer DINOv3 features fused into successive encoder stages, while color-frequency refinement (BFACG, SFFB) suppresses decoder-side chromatic collapse and skip-connection illumination leakage.
Experiments on CL3AN demonstrate that illumination-invariant semantic priors substantially outperform geometric and frequency-based alternatives, with controlled ablations confirming each component's contribution.
At the NTIRE 2026 ALN Challenge, CANDLE ranks 3rd on the Color Lighting track and achieves 2nd-place fidelity with the lowest FID on the White Lighting track, confirming generalization across both chromatic and luminance-dominant lighting conditions.


{
    \small
    \bibliographystyle{ieeenat_fullname}
    \bibliography{main}
}

\end{document}